\documentclass[10pt,twocolumn,letterpaper]{article}

\usepackage{iccv}              

\usepackage{times}
\usepackage{epsfig}
\usepackage{graphicx}
\usepackage{amsmath}
\usepackage{amssymb}
\usepackage{bm}
\usepackage{marvosym}
\usepackage{newtxtext}
\usepackage{newtxmath}

\definecolor{iccvblue}{rgb}{0.21,0.49,0.74}
\usepackage[pagebackref,breaklinks,colorlinks,allcolors=iccvblue]{hyperref}

\def\paperID{6451} 
\def\confName{ICCV}
\def\confYear{2025}

\title{RGE-GS: Reward-Guided Expansive Driving Scene Reconstruction \\
via Diffusion Priors}

\author{Sicong Du$^{1,\dag}$ \quad Jiarun Liu$^{1,2,\dag}$ \quad Qifeng Chen$^1$ \quad Hao-Xiang Chen$^3$ \quad Tai-Jiang Mu$^3$ \quad Sheng Yang$^{1,}$\textsuperscript{\Letter} \\
{\normalsize $^1$Unmanned Vehicle Dept., CaiNiao Inc., Alibaba Group \quad
$^2$Zhejiang University \quad
$^3$BNRist, Tsinghua University}\\
{\tt\small \{liujiarun.ljr,shengyang\}@cainiao.com, iedusicong@126.com}
}

\begin{document}
\maketitle

\begin{abstract}

A single-pass driving clip frequently results in incomplete scanning of the road structure, making reconstructed scene expanding a critical requirement for sensor simulators to effectively regress driving actions. Although contemporary 3D Gaussian Splatting (3DGS) techniques achieve remarkable reconstruction quality, their direct extension through the integration of diffusion priors often introduces cumulative physical inconsistencies and compromises training efficiency. To address these limitations, we present RGE-GS, a novel expansive reconstruction framework that synergizes diffusion-based generation with reward-guided Gaussian integration. The RGE-GS framework incorporates two key innovations: First, we propose a reward network that learns to identify and prioritize consistently generated patterns prior to reconstruction phases, thereby enabling selective retention of diffusion outputs for spatial stability. Second, during the reconstruction process, we devise a differentiated training strategy that automatically adjust Gaussian optimization progress according to scene converge metrics, which achieving better convergence than baseline methods. Extensive evaluations of publicly available datasets demonstrate that RGE-GS achieves state-of-the-art performance in reconstruction quality. Our source-code will be made publicly available at \url{https://github.com/CN-ADLab/RGE-GS}.

\end{abstract}

\renewcommand{\thefootnote}{}
\footnote{$\dag$: Equally contributed. \Letter: Corresponding author.}

\section{Introduction}
\label{sec:intro}

Autonomous driving simulators necessitate sophisticated street scene modeling that can effectively synthesize multi-view sensor data while adhering to realistic physical principles -- a fundamental capability critical for scalable data augmentation~\cite{zhang2025mapgs} and closed-loop simulation~\cite{li2024choose}.

Expansive scene modeling -- which involves performing physically plausible and perceptually realistic reconstruction on under-observed regions from single-pass real-world scans -- has become a pivotal capability for autonomous driving simulators to support reactive regressions.
While recent breakthroughs in controllable diffusion models~\cite{gao2023magicdrive,wang2024freevs} have achieved significant improvements of visual effects, these approaches remain limited in their adaptability for scene-specific training and tuning, as well as their computational efficiency for real-time Novel View Synthesis (NVS) along predefined trajectories.

Therefore, many state-of-the-art diffusion-aided reconstruction frameworks~\cite{wang2024drivedreamer,fan2024freesim,lu2024InfiniCube} incorporate generated scans into an explicit reconstruction pipeline to improve physical consistency, facilitate real-time photorealistic rendering, and enable interactive editing of reconstructed driving scenes. Nevertheless, integrating diffused scans with physical inconsistency into Gaussian reconstruction process may lead to unpredictable artifacts or compromised training efficiency. To date, existing frameworks have yet to adequately resolve this challenge, as they predominantly rely on reconstruction methods~\cite{Kerbl20233dgs,Huang20242DGS,Lu2024scaffoldgs} originally designed for real-world scans without necessary adaptations. 
Fine-tuning diffusion models on domain-specific data can help narrow the gap between generated images and real scans. However, this approach may still fall short in addressing inherent biases in the generated outputs. Without a robust mechanism to identify and reject these biases, the generated images may lack physical consistency, leading to potentially misleading results.

To more effectively leverage diffusion priors during expansive modeling, we propose RGE-GS, a reconstruction method that is compatible with recent advancements in controllable diffusion models~\cite{blattmann2023stable} and represents scenes using Gaussians. For generated scans, we quantify two pixel-level metrics revealing the reconstruction progress: (1) the coherence between newly generated diffusion priors and existing Gaussians and (2) the current reconstruction progress of under-observed and ongoing regions. 
As for the first metric, it can be used for quality inspection and data filtering for input diffusion priors, preserving mostly correct information for reconstruction. While fine-tuning diffusion models on domain-specific data or incorporating additional control information can help narrow the gap between generated images and real scans, these approaches alone may not fully address the challenges of continuous integration during reconstruction. To stress this issue, we propose a reward network which predicts the pixel-wise confidence of each generated image and assists in training the GS scene. Our experiments demonstrate that adversarially verifying the generated results as a complementary strategy is an effective means to reject biased outputs and enhance physical consistency.
The second metric identifies the under-reconstructed areas to focus on incompleted regions while preserving the already reconstructed parts. We design a differentiated training strategy that automatically applies different optimization techniques based on the convergence level of the Gaussians. This strategy not only enhances the overall learning performance of the Gaussian scene but also reduces training time by applying targeted optimization techniques for different regions.

In conclusion, RGE-GS makes the following key contributions:

\begin{itemize}
\item A novel reward-guided GS reconstruction method for expansive scene modeling via diffusion priors, achieving state-of-the-art quality and efficiency compared to previous diffusion-prior aided reconstruction methods.
\item A reward network to evaluate the pixel-level confidence of input diffusion priors considering the compatibility and progress of existing Gaussians.
\item A differentiated training strategy is proposed to adjust the optimization process of GS primitives based on different convergence, enhancing both efficiency and stability.
\end{itemize}
Extensive evaluations on publicly available datasets~\cite{ni2025paralane}  have demonstrated the effectiveness of our approach in leveraging these two metrics for adaptively handling diffusion priors, thereby improving both quality and efficiency during progressive reconstruction. 

\section{Related Work}

\subsection{Controllable Diffusion Models}

Controllable diffusion models for autonomous driving~\cite{li2023drivingdiffusion,gao2024vista} generate high-fidelity videos based on text prompts, reference sensor frames, and road structural information, including maps and obstacles. DriveDreamer~\cite{wang2024drivedreamer} aligns the above conditions to concatenate with noise images for diffusion, where trained temporal attention layers are leveraged to ensure the consistency of generated video frames. DriveDreamer2~\cite{zhao2024drivedreamer} further utilizes a fine-tuned large language model (LLM) to translate user prompts into agent trajectories and High Definition (HD) maps for controlling generation. MagicDrive~\cite{gao2023magicdrive} also encodes these conditions through tailored encoding strategies and further supports multi-sensor consistency. FreeVS~\cite{wang2024freevs} leverages projected colored LiDAR point cloud as pseudo-images for controlling the generation. Streetscapes~\cite{deng2024streetscapes} incorporates street and height maps for a layout-conditioned long-term generation.

In summary, these methods mainly focus on leveraging auxiliary information together with existing diffusion models. Since they use a unified big model for different driving clips to maintain temporal consistency, they also require a sufficient dataset and training process for this structural information.




\subsection{Diffusion-aided Reconstruction}

While controllable diffusion models propose to directly generate temporally coherent videos, their subsequent insights on leveraging reconstruction methods provide an applicable way of reducing both structural information and heavy models: DriveDreamer4D~\cite{Zhao2024drivedreamer4d} leverages diffusion priors via constructing temporally aligned cousin pairs between diffusion generated frames and Gaussian rendered frames. MagicDrive3D~\cite{gao2024magicdrive3d} uses consistent monocular depth prior, aligns exposure between cameras, and regards Gaussians as deformable for local dynamics to enhance the scene representation. FreeSim~\cite{fan2024freesim} proposes a progressive reconstruction strategy to add generated images from under-observed perspectives.

Although these methods with 3DGS (also see~\cite{yu2024sgd,han2024ggs,ni2024recondreamer,yan2024streetcrafter}) have successfully built appropriate loss functions leading reconstructed Gaussians compliant with diffusion priors, 3DGS -- as an explicit scene representation -- necessitates high cross-view geometry and appearance consistency, and unavoidably magnifies the challenge of convergence or results in conspicuous artifacts. Based on enhanced loss functions, our RGE-GS further guards the expansive reconstruction by examining the input diffusion priors based on the current state of Gaussians. Our proposed reward-guidance network does not require heavy parameters and training processes, and can be compatible with these reconstruction methods as an independent access-in module.

\begin{figure*}[t] 
\includegraphics[width=\linewidth]{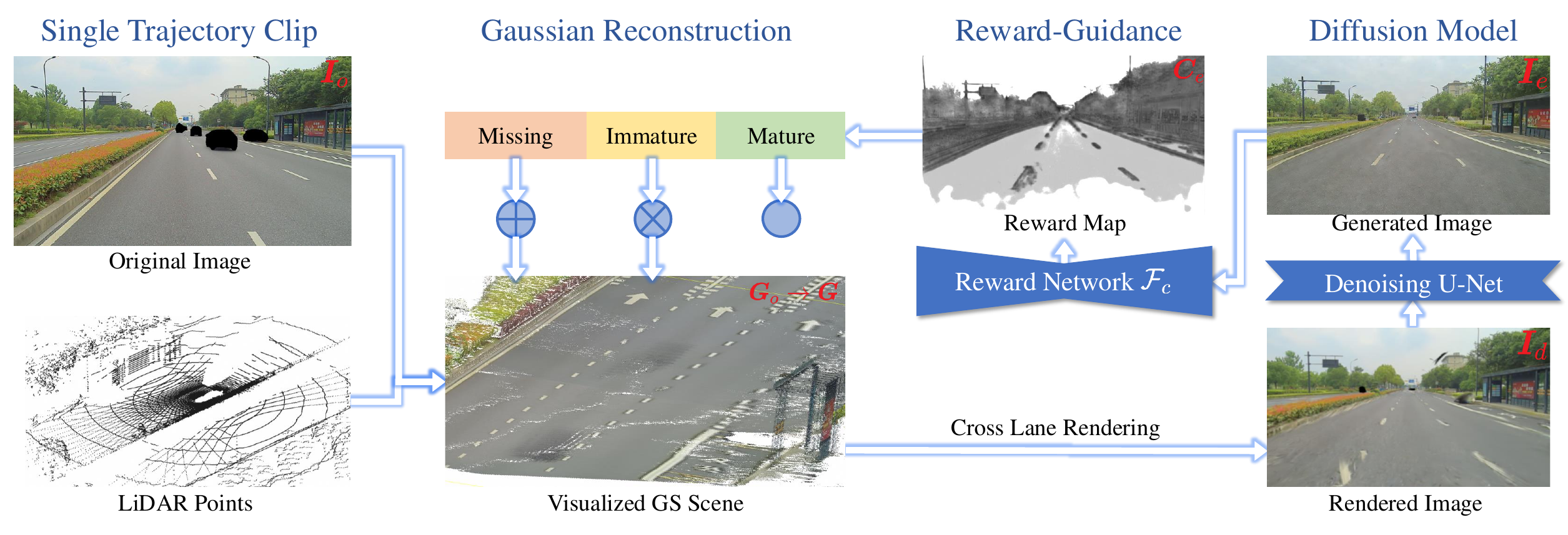}
\caption{The workflow of our proposed RGE-GS for expansive scene reconstruction. We use a single trajectory clip to reconstruct gaussians and feed rendered images to a diffusion model for novel view synthesis. Before sending diffused images back to the reconstruction, we design and train a reward guidance network to evaluate the consistency of generation and progress of reconstruction and infer a pixel-wise confidence during reconstruction.}
\label{fig:workflow} 
\end{figure*}

\subsection{Indicating Gaussian Reconstruction Progress}

GS-Planner~\cite{jin2024gsplanner} represents the completeness of partially reconstructed Gaussians for planning the active scanning trajectory w.r.t. the information gain. 3DGS-Enhancer~\cite{liu20243dgsenhancer} proposes to compute image and pixel levels of confidence through perspective parallax and Gaussians' covariance, respectively. ReconX~\cite{liu2024reconx} focuses on a global consistency on a series of generated frames to simultaneously obtain multiple confidence maps through the training process of DUSt3R. InstantSplat~\cite{fan2024instantsplat} introduces a confidence-aware per-point optimizer designed to facilitate convergence during self-supervised optimization. GaussianPro~\cite{cheng2024gaussianpro} proposes a progressive propagation strategy using geometrical values from PatchMatch~\cite{barnes2009patchmatch} to indicate under-reconstruction regions.
Among methods using Neural Radiance Field (NeRF) Representation~\cite{mildenhall2021nerf}, NeRF On-the-go~\cite{ren2024nerf} uses uncertainty from MLP to train models. Our method combines progress and consistency metrics through a unified model and proposes strategic improvements to better reconstruct scene regions at different completeness levels.

\section{Methodology of RGE-GS}
\label{sec:method}

\subsection{Preliminaries}
\label{sec:method:preliminary}


\textbf{3D Gaussian Splatting.} 3DGS~\cite{Kerbl20233dgs} characterizes the scene through a collection of anisotropic Gaussians defined within the 3D space. Each Gaussian \( \boldsymbol{G} \) is assigned several parameters: opacity \(\boldsymbol o \in \mathbb{R} \), spherical harmonics (SH) coefficients \(\boldsymbol z \in \mathbb{R}^k \), a position vector \( \boldsymbol\mu \in \mathbb{R}^3 \), a rotation quaternion \(\boldsymbol q \in \mathbb{R}^4 \), and a scale factor \(\boldsymbol s \in \mathbb{R}^3 \). The Gaussian kernel distribution is expressed as:
\begin{equation}
\boldsymbol{G}(\mathbf{x}) = \exp(-\frac{1}{2}(\mathbf{x} - \boldsymbol{\mu})^{\top} \mathbf{\Sigma}^{-1}(\mathbf{x} - \boldsymbol{\mu})),
\end{equation}
where the covariance matrix is defined as \( \boldsymbol{\Sigma} = \mathbf{R} \mathbf{S} \mathbf{S^{\top}} \mathbf{R^{\top}} \), with \( \mathbf{S} \) representing the scaling matrix dictated by \( \boldsymbol{s} \) and \( \mathbf{R} \) being the rotation matrix derived from \( \boldsymbol{q} \). Given the camera's extrinsic parameters \( \mathbf{W} \), the 2D covariance matrix in screen space is computed as follows:

\begin{equation}
\boldsymbol{\Sigma^*} = \mathbf{J W} \boldsymbol\Sigma \mathbf{W^{\top} J^{\top}},
\end{equation}

\textbf{Latent Diffusion Models.} Latent diffusion models (LDMs)~\cite{rombach2022high} represent a category of generative models that effectively learn the distribution of the latent variable $\mathbf{X}$ derived from the dataset ${\mathbf{D}}$. These models work in conjunction with an autoencoder (or tokenizer) that transforms the data into the latent space defined as $\mathbf{X}=\epsilon(\mathbf{D})$. A diffusion model initiates the process with a noise vector $\mathcal{N}(0, \mathbf{I})$ and progressively denoises it to produce a sample $\mathbf{X}$, which can be influenced by a condition ${\mathbf{C}}$. Subsequently, the decoder is employed on the denoised latent representation to create the output data $\mathbf{\hat{D}}$. LDMs have demonstrated their effectiveness and efficiency in capturing various data modalities, such as images~\cite{rombach2022high} and videos~\cite{blattmann2023stable}.

\subsection{Framework Overview}
\label{sec:method:overview}
The overall framework of our RGE-GS is shown in Fig.\ref{fig:workflow}. The first round, we use the ground truth images $\bigcup { \boldsymbol{I}_o }$ from the original viewpoints to reconstruct the GS primitives, denoted as $\boldsymbol{G}_o$. To prepare for the expansion of GS, we shift to extrapolated free viewpoints along the original sensor trajectory to render degraded images $\bigcup{ \boldsymbol{I}_d }$ ,and feed these into a diffusion model to generate 
pseudo images $\bigcup{ \boldsymbol{I}_e }$.
The Second round, $\bigcup{ \boldsymbol{I}_e }$ and $\bigcup\boldsymbol{I}_o$ are used as input to  retrain $\boldsymbol{G}_o$. Especially, we introduce a reward network $\mathcal{F}_c$ (Sec.~\ref{sec:method:reward}) to identifies high-demand and high-confidence regions corresponding to each pixel of generated image. It is jointly optimized with $\boldsymbol{G}_o$, and the output of $\mathcal{F}_c$, reward map $\bigcup{\boldsymbol{C}_e }$, served as an adversarial verification to assist the consistency of scene expansion. Meanwhile, we enhance the expansive reconstruction process with such auxiliary info by differentiating the initialization and training strategies and obtain the final scene representation $\boldsymbol{G}$ (Sec.~\ref{sec:method:recon}).

\subsection{Reward Network}
\label{sec:method:reward}

\begin{figure}[tb] 
\includegraphics[width=\linewidth]{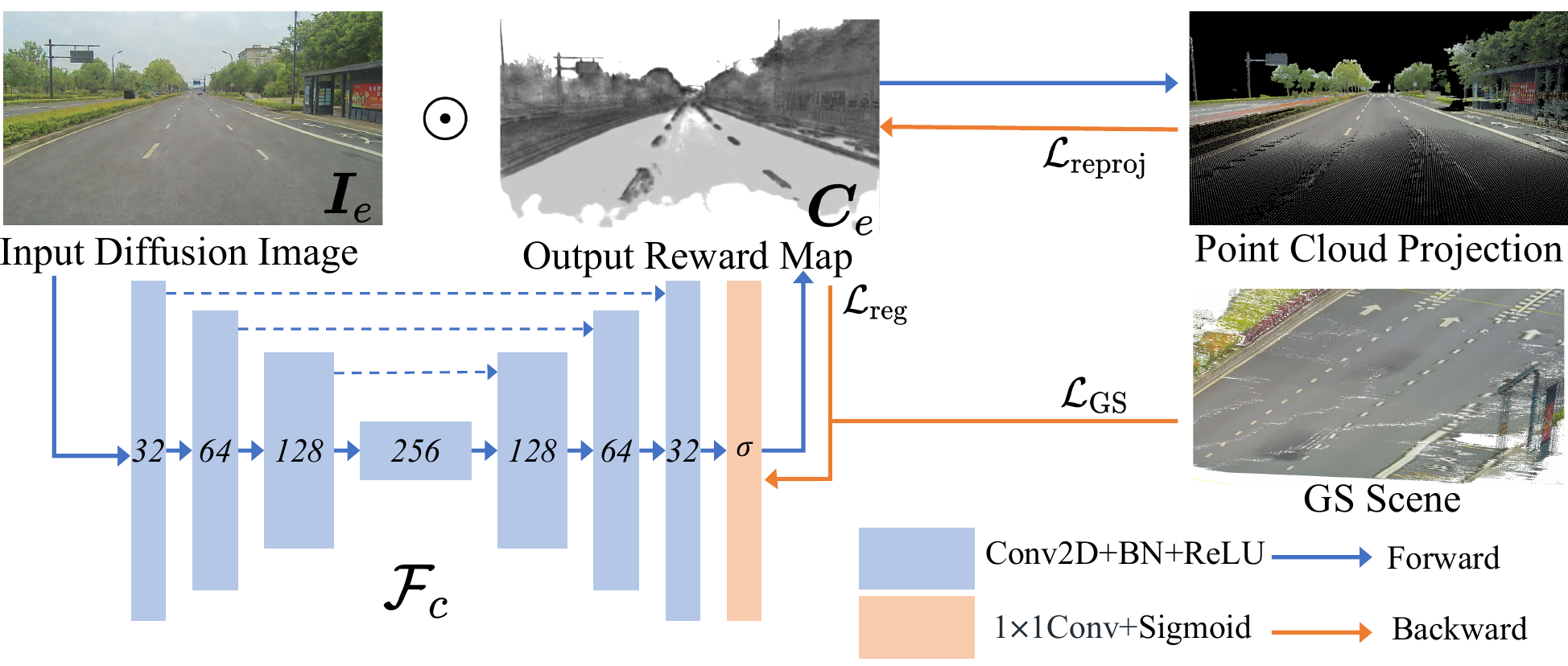}
\caption{The illustration of reward network.}
\label{fig:confnet} 
\end{figure}

Our framework systematically integrates the prior images generated by the diffusion model into GS training through a lightweight reward network. This auxiliary module is trained in an adversarial strategy to produce pixel-wise reward scores $\boldsymbol{C}_e$ that dynamically modulate the contribution of generated priors.

\textbf{Network structure.} As shown in Fig.~\ref{fig:confnet}, the reward network $\mathcal{F}_c:\boldsymbol{I}_e\in\mathbb{R}^\mathrm{H\times W\times3}\rightarrow\boldsymbol{C}_e\in\mathbb{R}^\mathrm{H\times W}$ adopts a lightweight U-Net architecture with encoder-decoder symmetry and residual skip connections~\cite{unet2015}. Comprising three downsampling and upsampling stages, the network processes input images through successive convolution blocks to extract multi-scale features, followed by transposed convolutions for spatial recovery. Skip connections bridge encoder and decoder features to preserve structural details, culminating in a prediction head with sigmoid activation that outputs pixel-wise confidence scores between 0 and 1.

\textbf{Training strategy.} Given the absence of ground truth for confidence values, the reward network is constructed as an adversarial verification module, facilitating unsupervised learning during joint training with the GS reconstruction framework. This design allows for the auxiliary generation of confidence maps.
Specifically, for each scene, we first integrate the reward network into the GS pipeline and train it for multiple iterations. During this process, the reward network is guided by the following supervisions: (1) a reprojection error loss $\mathcal{L}_{\textrm{reproj}}$ that enforces consistency between point cloud projections and diffusion priors, and (2) a binarization regularization term $\mathcal{L}_{\textrm{reg}}$ that prevents forming indistinguishable reward maps. 

The reprojection error loss is defined as Eq.~\ref{eq:reproj_loss}:
\begin{equation}
\mathcal{L}_{\textrm{reproj}}=\frac{1}{\mathrm{H \cdot W}}\|\boldsymbol{C}_e\odot(\pi(\mathbf{P},\mathbf{T}_e,\mathbf{K})-\boldsymbol{I}_e)\|_2,
\label{eq:reproj_loss}
\end{equation}
where $\mathbf{P}$ is the colored LiDAR point cloud generated by recolor each LiDAR points with their positions in the world reference frame and their visible colors. $\pi(\mathbf{P},\mathbf{T},\mathbf{K})$ represents the projection function to transpose 3D points to 2D image with camera pose $\mathbf{T}$ and intrinsic $\mathbf{K}$. The reprojection of colored point cloud provides coarse explicit spervision to the reward map, which facilitates the gradient optimization of the reward network during adversarial training.

The binarization regularization term in Eq.~\ref{eq:reg_loss} is designed to encourage the confidence distribution to concentrate around 0 or 1, thereby avoiding ambiguous predictions. 
\begin{equation}
\mathcal{L}_{\textrm{reg}}=\frac{\boldsymbol{C}_e \cdot (1-\boldsymbol{C}_e)}{\mathrm{H\cdot W}}.
\label{eq:reg_loss}
\end{equation}


The overall training loss of reward network $\mathcal{L}_{\textrm{reward}}$ is defined as Eq.~\ref{eq:reward_loss}: 
\begin{equation}
\mathcal{L}_{\textrm{reward}}=\lambda_{\text{reproj}}\mathcal{L}_{\textrm{reproj}}+\lambda_{\text{reg}}\mathcal{L}_{\textrm{reg}}+\mathcal{L}_{\textrm{GS}},
\label{eq:reward_loss}
\end{equation}

where $\lambda_{\textrm{reproj}}=0.5$ and $\lambda_{\textrm{reg}}=0.3$. $\mathcal{L}_{\textrm{GS}}$ represents the total loss of GS reconstruction, which will be discussed further in Sec.~\ref{sec:method:recon}. The output of reward network $\boldsymbol{C}_e$ is applied to corresponding $\boldsymbol{I}_e$ during training. 

\subsection{Differentiated Training Strategy}
\label{sec:method:recon}

After $\boldsymbol{G}_o$, $\bigcup{\boldsymbol{I}_e}$ (Sec.~\ref{sec:method:preliminary}) and $\bigcup{\boldsymbol{C}_e}$ (Sec.~\ref{sec:method:reward}) are obtained,
we categorize the GS primitives into three groups and apply different training strategies based on the convergence status.


\textbf{1. Missing Gaussians.} 
For regions visible in the novel views $\bigcup{\boldsymbol{I}_e}$ but occluded in $\bigcup{\boldsymbol{I}_o}$, 
we adopt the same initialization strategy used in the first reconstruction pass and employ the following two methods: (1) for areas covered by stitched LiDAR point clouds, we initialize the positions of the Gaussians based on the geometric distribution of the LiDAR points, and (2) for regions inadequately scanned by LiDAR sensors, we utilize depth estimation from DepthAnything-V2~\cite{yang2024depth} and calibrate the scale by minimizing the Euclidean distance relative to existing LiDAR measurements. This latter strategy enables us to successfully initialize a significant number of unobserved regions, thereby alleviating the time-consuming and unstable aspects of multi-view geometry during Gaussian training.


\textbf{2. Immature Gaussians.}
Incomplete scans would result in Gaussians with large view-space positional gradients when observed on novel view images $\boldsymbol{I}_e$ since they cannot be sufficiently trained due to limited multi-view parallax.
Intuitively, this phenomenon is because some regions are not yet fully reconstructed, and the optimization process tries to address this by moving Gaussians. Consequently, we identify these immature primitives using a predefined gradient threshold of $\lambda=5\times 10^{-4}$ for further optimization until convergence with extra supervision from $\boldsymbol{I}_e$.


\textbf{3. Mature Gaussians.}
For the rest of the existing Gaussians regarded as matured, they no longer need to be affected by generative scans, so we fix their attributes to ensure the stability of reconstruction, which also accelerates the training procedure.


The reward maps $\bigcup{\boldsymbol{C}_e}$ are then applied to the corresponding diffusion prior for GS reconstruction. The reconstruction loss function is defined as:
\begin{equation}
\begin{aligned}
    \mathcal{L}_{\boldsymbol{I}_e} &= \lambda_{\textrm{RGB}}\mathcal{L}_{RGB}(\boldsymbol{I}_d,\boldsymbol{I}_e,\boldsymbol{C}_e) \\
    &+ (1-\lambda_{\textrm{RGB}})\mathcal{L}_{\textrm{SSIM}}(\boldsymbol{I}_d,\boldsymbol{I}_e,\boldsymbol{C}_e) + \lambda_{\boldsymbol{I}_e}\mathcal{L}_{\textrm{LPIPS}}(\boldsymbol{I}_d,\boldsymbol{I}_e),
\end{aligned}
\end{equation}
where $\mathcal{L}_{\textrm{RGB}}=0.8$ is the L1 norm with the weighted reward map $\bigcup{\boldsymbol{C}_e}$ multiplied on the gaussian rendering $\boldsymbol{I}_d$ and generated image $\bigcup{ \boldsymbol{I}_e }$ respectively. $\mathcal{L}_{\textrm{ssim}}$ is the structural similarity index measure~\cite{wang2004image}. $\mathcal{L}_{\textrm{LPIPS}}$ is the perceptual similarity metric\cite{zhang2018lpip} and $\lambda_{\boldsymbol{I}_e}$ is set to 0.01.

The definition of loss function during the GS reconstruction process on differentiated training strategy is as follows,
\begin{equation}
\begin{aligned}
\mathcal{L}_{\textrm{GS}} &=  \mathcal{L}_{\boldsymbol{I}_o}+\mathcal{L}_{\boldsymbol{I}_e}, \\
    \mathcal{L}_{\boldsymbol{I}_o} &= \lambda_{\boldsymbol{I}_o}\mathcal{L}_{1}(\boldsymbol{I},\boldsymbol{I}_o)+ (1-\lambda_{\boldsymbol{I}_o})\mathcal{L}_{\textrm{SSIM}}(\boldsymbol{I},\boldsymbol{I}_o),
\end{aligned}
\end{equation}
where $\lambda_{\boldsymbol{I}_o}=0.8$, $\boldsymbol{I}$ represents the ground truth image of the original lane.

\section{Evalution}
\label{sec:exp}

\begin{figure*}[t] 
\includegraphics[width=\linewidth]{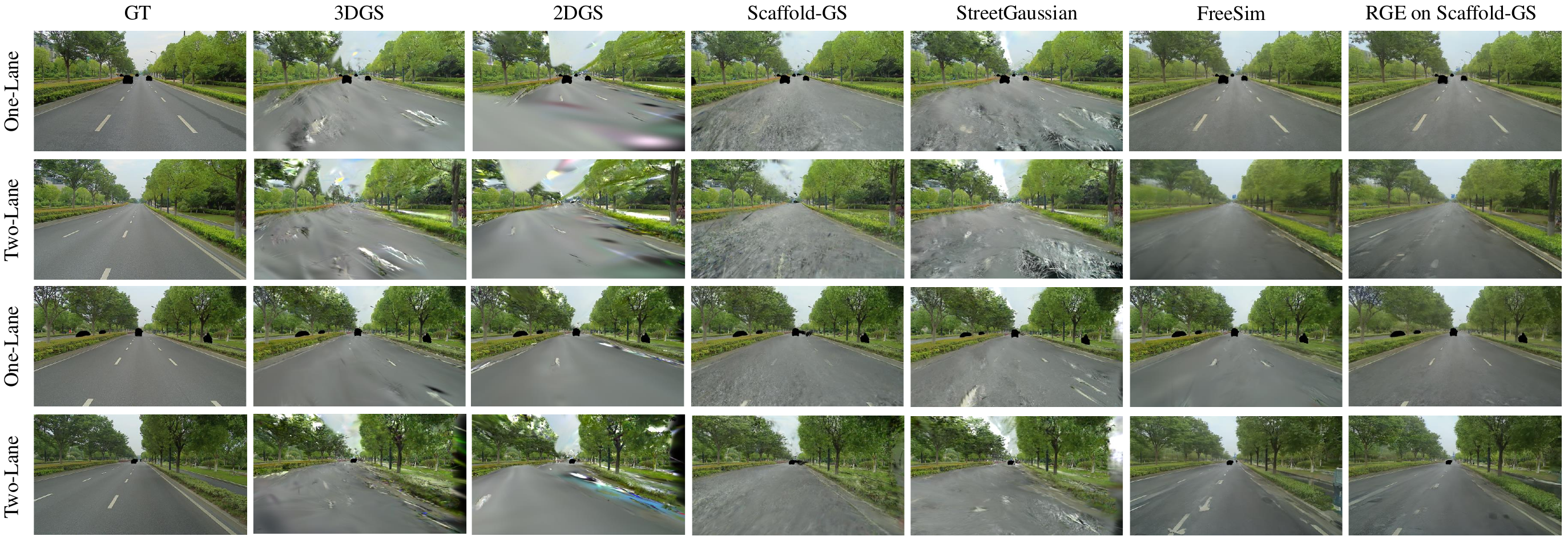}
\caption{Comparisons for cross-lane view synthesis with publicly available methods, demonstrating our effectiveness of using a reward network and differentiated training strategy for expansive reconstruction. Please refer to our \textbf{supplementary video} for more examples.}
\label{fig:methods-compare} 
\end{figure*}

\subsection{Implementation and Experimental Settings}
\label{sec:exp:impl}

\textbf{Dataset.} We conduct experiments on Para-Lane~\cite{ni2025paralane} datset and Waymo Open Datset. On the Para-Lane dataset, we select four scenes containing three lanes (left, middle, right), comprising 250 front-view images of each lane scanned at 1280×720@10Hz to evaluate the aforementioned method with publicly available baselines


\textbf{Implementation details of the diffusion model.}
Our generative model is initialized using the pre-trained weights from SD v1.5~\cite{rombach2022high}. We train the parameters of U-Net~\cite{ronneberger2015u} and ControlNet~\cite{zhang2023adding} for 50,000 iterations across 4 A100 GPUs with a batch size of 8 per GPU while keeping the VAE~\cite{kingma2013auto} parameters frozen. The training dataset is constructed from the Para-Lane dataset using method in ~\cite{wang2024freevs}, containing about 60,000 pairs. During training, the learning rate is set to 5e-5 and the resolution of generated images remains the same with input real-scans.

\textbf{Implementation details of the reward network.}
The reward network is initialized with random parameters, except for a bias constraint applied before the prediction head to keep its output away from all-zero. The network employs the AdamW optimizer with standard parameter settings. The learning rate is initialized at 5e-4, undergoes linear decay during the first 1,000 iterations, and then follows a cosine annealing schedule. The network is jointly trained with Gaussian scenarios while synchronously updating parameters, undergoing a total of 5,000 iterations. After that, the reward network is frozen and we only use it to generate reward output $\bigcup\boldsymbol{C}_e$ for the following training process.

\textbf{Implementation details of the Gaussian reconstruction.}
Our proposed method is compatible for most publicly available Gaussian reconstruction frameworks. In our ablation (also see Tab.~\ref{tab:nvs_comparisions}), we select 3DGS~\cite{Kerbl20233dgs} and Scaffold-GS~\cite{Lu2024scaffoldgs} as our baselines to demonstrate our adapt ability. The Gaussian densification process begins and ends at 500 and 15,000 iterations, respectively, with a gradient threshold for mature primitive selection set at 4e-4. The number of initialization points is configured to 50,000.

\textbf{Repeat of publicly available methods.}
\setcounter{footnote}{0}
\renewcommand{\thefootnote}{\arabic{footnote}}
We replicate the current state-of-the-art cross-lane view synthesis methods FreeSim~\cite{fan2024freesim} and ReconDreamer~\cite{ni2024recondreamer} since both are not completely open-sourced as of now\footnote{The currently open-sourced version of ReconDreamer does not include the generation component, mentioned as DriveRestorer in its draft.}. Specifically, our training framework is built upon a dynamic-static decoupled GS reconstruction pipeline, where the generation results from our proposed diffusion model serve as cross-lane novel view priors. To simplify the training process and directly get the upper-bound performance of the baseline method, we directly use generated images of cross-lane views for training, bypassing the progressive view generation approach that transitions from near to far.

\begin{table}[htbp]
\caption{Fidelity of novel view synthesis with different methods on Para-Lane dataset. Asterisks stand for our reimplementation.}
\centering
\fontsize{8pt}{9.6pt}\selectfont
\setlength{\tabcolsep}{1.5pt}
\begin{tabular}{l|ccc|ccc}
\toprule
\toprule
Method                       & \multicolumn{3}{c|}{One-Lane Expansion} & \multicolumn{3}{c}{Two-Lane Expansion} \\
Metrics                      & PSNR$\uparrow$ & SSIM$\uparrow$ & LPIPS$\downarrow$ & PSNR$\uparrow$ & SSIM$\uparrow$ & LPIPS$\downarrow$    \\
\midrule
\midrule
3DGS~\cite{Kerbl20233dgs}             & 16.12 & 0.519 & 0.461 & 14.50 & 0.462 & 0.526 \\
2DGS~\cite{Huang20242DGS}             & 15.04 & 0.527 & 0.476 & 13.04 & 0.483 & 0.529 \\
Street Gaussians~\cite{yan2024street} & 16.76 & 0.517 & 0.450 & 16.26 & 0.485 & 0.506\\
Scaffold-GS~\cite{Lu2024scaffoldgs}   & 17.25 & 0.519 & 0.453 & 16.70 & 0.471 & 0.511\\
FreeSim*~\cite{fan2024freesim}        & 19.07 & 0.537 & 0.424 & 17.96 & 0.514 & \textbf{0.466} \\
\midrule
RGE on 3DGS                        & {18.73} & {0.556} & {0.461} & {18.01} & \textbf{0.542} & {0.483} \\
RGE on Scaffold-GS                 & \textbf{19.72} & \textbf{0.586} & \textbf{0.415} & \textbf{18.11} & {0.519} & {0.468} \\
\bottomrule
\bottomrule
\end{tabular}
\label{tab:nvs_comparisions}
\end{table}

\subsection{Main Results}

\textbf{Quantitative comparison.}
On the Para-Lane Dataset, we present the performance of novel view synthesis for scene expansion across one lane and two lanes in Table~\ref{tab:nvs_comparisions}, with lateral trajectory shifts of approximately 3.5m and 7m, respectively.
In both comparisons, our proposed RGE strategy significantly outperforms previous reconstruction-based methods. 
This improvement is primarily due to the incorporation of a confidence-aware diffusion prior during the reconstruction process for the novel views. 
Additionally, our method exhibits a slight performance drop in two-lane expansion compared to one-lane expansion, which can be attributed to the less effective guidance information provided to the diffusion model from the rendering results when larger lateral shifts are involved.
Furthermore, RGE, when applied to both 3DGS and Scaffold-GS, demonstrates a consistent trend, highlighting the adaptability of our method.

On the Waymo dataset, we select 10 scenes that offer relatively spacious movement areas, ensuring that camera movements do not intrude into the obstacle regions to accommodate the trajectory shifting.
Due to the cross-lane ground truth images being unavailable in the Waymo dataset, we adopt the FID score based on InceptionV3 for off-trajectory evaluation following ~\cite{ni2024recondreamer}, with different lateral trajectory shifts from 1m to 5m. As a result shown in Tab.~\ref{tab:nvs_waymo_comparisions}, our proposed method outperforms the previous methods, especially when the lateral shift distance is relatively large. This is because the degradation of GS rendering quality due to large lateral movements has a cascading effect on the diffusion output, making the reward network mechanism increasingly critical for reconstruction in such scenarios.

\begin{table}[htbp]
\caption{Fidelity of novel view synthesis with different methods on Waymo dataset. Asterisks stand for our reimplementation.}
\centering
\fontsize{8pt}{10pt}\selectfont
\setlength{\tabcolsep}{5.7pt}
\begin{tabular}{l|ccccc}
\toprule
\toprule
Method                        & \multicolumn{5}{c}{Lane Shift views (FID $\downarrow$)} \\
Metrics                       & @1m & @2m & @3m & @4m & @5m \\
\midrule
\midrule
Scaffold-GS                   & 62.93 & 84.32 & 124.57 & 148.32 & 189.25  \\
FreeSim*                      & 56.87 & 79.23 & 111.77 & 138.07 & 174.12  \\
ReconDreamer*                 & 59.76 & 76.44 & 117.46 & 142.67 & 167.97  \\
RGE on Scaffold-GS            & \textbf{52.41} & \textbf{71.56} & \textbf{102.37} & \textbf{117.08} & \textbf{129.12}  \\
\bottomrule
\bottomrule
\end{tabular}
\label{tab:nvs_waymo_comparisions}
\end{table}

\textbf{Qualitative comparison.}  Fig.~\ref{fig:methods-compare} shows the visual comparisons between RGE on Scaffold-GS and other methods in novel views on Para-Lane dataset. We can find that images rendered by the baseline algorithms exhibit are filled with speckles and ghosting. Our method, however, significantly improves the rendering quality. 

\subsection{Ablation Results}

\begin{figure*}[ht]
    \centering
    \includegraphics[width=\linewidth]{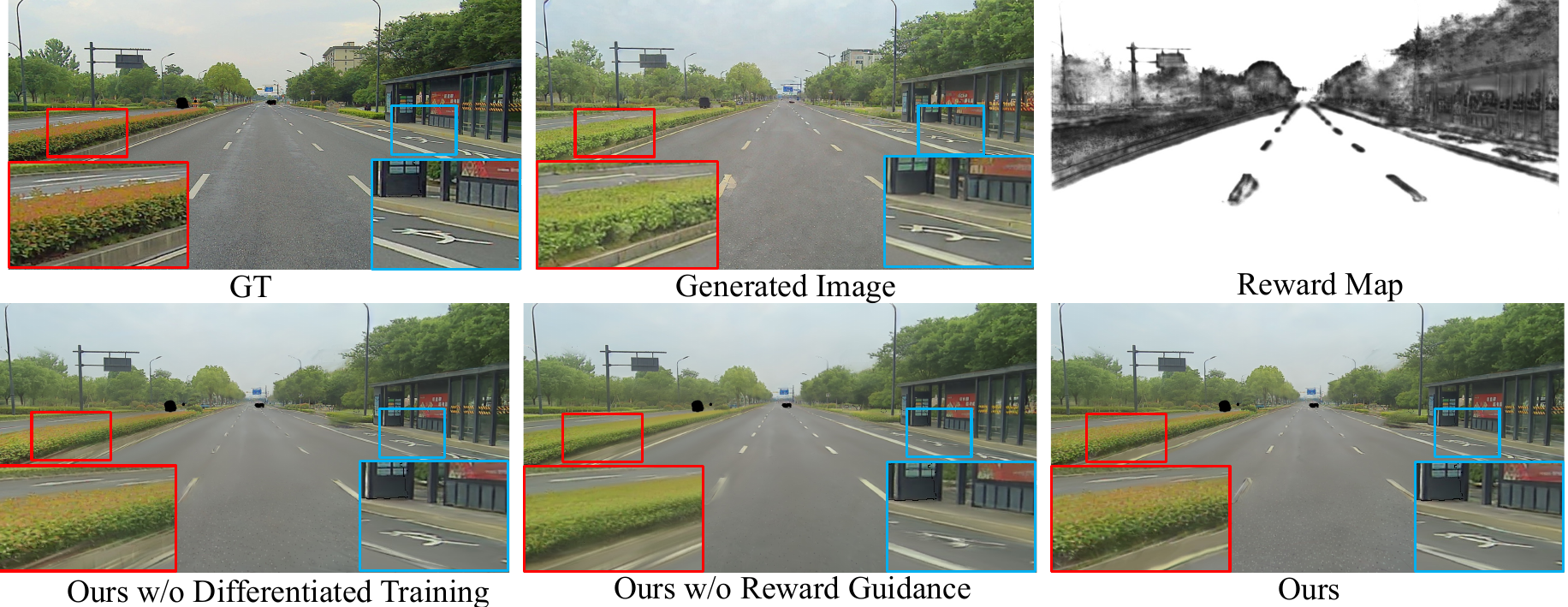}
    \caption{Ablations of cross-lane view synthesis results with different experiment settings. Notice the restoration of vegetation color and the expression of high-frequency details.}
    \label{fig:ablation}
\end{figure*}

\begin{figure*}[ht]
    \centering
    \includegraphics[width=\linewidth]{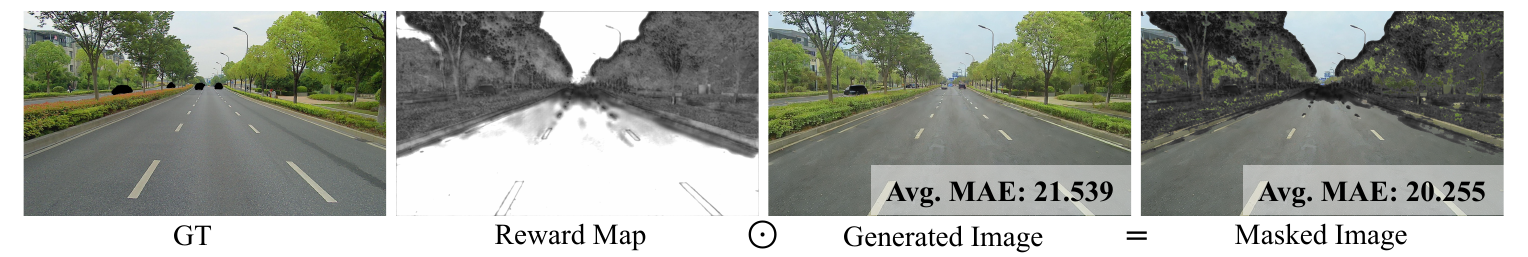}
    \caption{Visualization of reward map and layered diffusion output. The darker the areas in the reward map, the lower the confidence.}
    \label{fig:rebuttal}
\end{figure*}

\textbf{Ablations on the reward network.}
We use Scaffold-GS as our baseline method in ablations.
Tab.~\ref{tab:ablation} and Fig.~\ref{fig:ablation} demonstrates the significance of the reward inference module outputting $\boldsymbol{C}_e$. The experimental results demonstrate that our reward net effectively constrains the diffusion outputs through adaptive regulation, thus mitigating potential distortions in diffusion priors caused by implausible generation artifacts while preserving critical anatomical authenticity. Notice that although classifying Gaussian primitives can improve the effectiveness of training under pseudo views, the supervision of training comes directly from the diffusion output, which risks propagating error information. The reward network can limit these error priors at the source, ensuring stable training. We further validate the correctness of the network output by overlaying the reward maps on diffusion outputs and measuring MAE (Mean Absolute Error) of valid pixels to ground truth images on Para-Lane dataset. As shown in Fig.~\ref{fig:rebuttal}, the reward map effectively limits the adverse effects of erroneous outputs, confirming our training strategy’s effectiveness.

\begin{table}[htbp]
\caption{Ablation study on novel view synthesis with different pipeline settings on Para-Lane dataset with one-lane expansion.}
\centering
\fontsize{8pt}{9.6pt}\selectfont
\setlength{\tabcolsep}{3pt}
\begin{tabular}{l|ccc|c}
\toprule
\toprule
Metrics                      & PSNR$\uparrow$ & SSIM$\uparrow$ & LPIPS$\downarrow$ & Training (min) $\downarrow$ \\
\midrule
\midrule
w/o RewardNet               & 19.24 & 0.544 & 0.427 & 91 \\
w/o Diff-Train              & 19.55 & 0.552 & 0.419 & 112 \\
RGE on Scaffold-GS          & \textbf{19.72} & \textbf{0.586} & \textbf{0.415} & \textbf{89} \\
\bottomrule
\bottomrule
\end{tabular}
\label{tab:ablation}
\end{table}

\begin{table}[htbp]
\caption{Ablation study on novel view synthesis with different diffusion prior models on ParaLane dataset.}
\centering
\fontsize{8pt}{9.6pt}\selectfont
\setlength{\tabcolsep}{0.5pt}
\begin{tabular}{l|ccc|ccc}
\toprule
\toprule
Method                       & \multicolumn{3}{c|}{One-Lane Expansion} & \multicolumn{3}{c}{Two-Lane Expansion} \\
Metrics                      & PSNR$\uparrow$ & SSIM$\uparrow$ & LPIPS$\downarrow$ & PSNR$\uparrow$ & SSIM$\uparrow$ & LPIPS$\downarrow$    \\
\midrule 
\midrule
RGE\_SD on Scaffold-GS       & {19.72} & {0.586} & {0.415} & {18.11} & {0.519} & {0.468} \\
SVD on Scaffold-GS       & {19.52} & {0.552} & {0.422} & {18.08} & {0.499} & {0.474} \\
RGE\_SVD on Scaffold-GS      & \textbf{19.95} & \textbf{0.601} & \textbf{0.404} & \textbf{18.63} & \textbf{0.520} & \textbf{0.432} \\
\bottomrule
\bottomrule
\end{tabular}
\label{tab:diffusion_study}
\end{table}

\textbf{Ablations on the differentiated training strategy.} Tab.~\ref{tab:ablation} also showcase the quantitative results of the ablation study on differentiated training strategy. The results indicate the effectiveness of our proposed graduated training method. Specifically, the training time is reduced with differentiated training strategy because the number of Gaussians requiring gradient computation is reduced.

\textbf{Additional comparisons on different diffusion prior models.} We conduct an ablation study with two fine-tuned diffusion models on the ParaLane dataset: stable diffusion v1.5 (SD) ~\cite{rombach2022high} and stable video diffusion (SVD) ~\cite{blattmann2023stable}. As shown in Tab.~\ref{tab:diffusion_study}, RGE\_SVD results in light improvements compared to RGE\_SD due to its enhanced temporal consistency, and the proposed RGE strategy can also enhance the performance of SVD. The ablation study proves that the reward map along with the differentiable training strategy can reduce the impact of hallucinations generated by different diffusion prior source.

\section{Conclusion and Future Work}
In this paper, we introduced RGE-GS, a novel method for expansive scene reconstruction that synergizes diffusion-based generation with reward-guided Gaussian integration. Our approach addresses the challenges of incomplete scene reconstruction from single-pass real scans by leveraging diffusion priors in a controlled and efficient manner. The key contributions of RGE-GS include a reward network that evaluates the pixel-level confidence of input diffusion priors, and a differentiated training strategy that categorizes Gaussians into missing, immature, and mature for optimized reconstruction. Extensive evaluations on publicly available datasets demonstrate that RGE-GS achieves state-of-the-art performance in both reconstruction quality and training efficiency.

Our future work includes adopting video generation
models to enchance the temporal consistency in diffusion model outputs, and exploring the integration of the reward network in a feed-forward Gaussian Splatting reconstruction pipeline for real-time reconstruction.

\section*{Acknowledgements}

We thank the reviewers for the valuable discussions. This research was supported by the Zhejiang Provincial Natural Science Foundation of China under Grant No. LD24F030001.

{\small
\bibliographystyle{ieeenat_fullname}
\bibliography{main}
}

\end{document}